\pdfoutput=1

\documentclass[11pt]{article}

\usepackage[final]{EMNLP2023}

\usepackage{times}
\usepackage{latexsym}
\usepackage{enumitem}
\usepackage{subcaption}
\usepackage[]{multirow}

\usepackage{graphicx}

\usepackage[T1]{fontenc}

\usepackage[utf8]{inputenc}

\usepackage{microtype}

\usepackage{inconsolata}
\usepackage{xcolor}
\usepackage{soul} 

\usepackage{xspace}
\usepackage{pifont}

\newcommand{\camReady}[1]{\textcolor{black}{#1}}

\newcommand{\tool}{FLEEK\xspace}
\newcommand{\ignore}[1]{}
\newcommand{\llm}{GPT-3\xspace}

\newcommand{\highlight}[2][white]{\sethlcolor{#1}\hl{#2}}

\usepackage{tcolorbox}

\title{\highlight{FLEEK: Factual Error Detection and Correction with Evidence Retrieved from External Knowledge}}

\author{Farima Fatahi Bayat$^1$\thanks{~~Work done while the author was an intern at Apple.}, Kun Qian$^2$, Benjamin Han$^2$, Yisi Sang$^2$, Anton Belyi$^2$, \\ {\bf Samira Khorshidi$^2$, Fei Wu$^2$, Ihab F. Ilyas$^2$, Yunyao Li$^2$} \\
  University of Michigan - Ann Arbor$^1$, Apple$^2$ \\
  farimaf@umich.edu, {\{kunqian, ben.b.han, yisi\_sang, a\_belyy\}@apple.com} \\
  {\{samiraa, fwu7, iilyas, yunyaoli\}@apple.com}
}
  
\begin{document}

\maketitle
\begin{abstract}
Detecting factual errors in textual information, whether generated by large language models (LLM) or curated by humans, is crucial for making informed decisions. LLMs' inability to attribute their claims to external knowledge and their tendency to hallucinate makes it difficult to rely on their responses. Humans, too, are prone to factual errors in their writing. Since manual detection and correction of factual errors is labor-intensive,  developing an automatic approach can greatly reduce human effort. We present \tool, a prototype tool that automatically extracts factual claims from text, gathers evidence from external knowledge sources, evaluates the factuality of each claim, and suggests revisions for identified errors using the collected evidence. Initial empirical evaluation on fact error detection (77-85\% F1) shows the potential of \tool. A video demo of FLEEK can be found at \url{https://youtu.be/NapJFUlkPdQ}.

 

\end{abstract}

\section{Introduction}

While textual information offers a convenient and efficient means of communication, it is critical to acknowledge its potential for misuse or unintended consequences. False or misleading information spreads easily over online platforms \cite{10.1145/2874239.2874267}. Additionally, the emergence of powerful Large Language Models (LLMs) such as GPT models \footnote{\url{https://platform.openai.com/docs/models}}, Vicuna \cite{vicuna2023}, and Alpaca \cite{alpaca} \camReady{have} introduced a new avenue for knowledge-seeking inquiries. These models, however, have a tendency to hallucinate and provide creative and fluent responses that are not factually accurate \cite{pan2023unifying}. The limitation of LLMs to attribute their responses to external valid evidence makes it challenging to trust their responses. Therefore, having a robust fact-checking mechanism is of paramount importance to ensure the integrity and accuracy of information. 

Previous works \cite{DBLP:journals/corr/abs-1909-03745, diliello2022paragraphbased, liu-etal-2020-fine} and systems like FACTGPT \footnote{\url{https://factgpt-fe.vercel.app/}}, typically formulates the fact verification as a classification task where the input consists of the evidence sentence(s) and the claim, and the output is a label indicating the veracity of the entire claim as SUPPORTED, REFUTED, or IRRELEVANT. \ignore{These labels do not provide detailed explanations, highlight the claim spans that are supported or refuted, nor do they connect those spans to the relevant evidence.} As a concrete example, if the claim is ``{\em United States is in North America and has 51 states}'', then a sentence-level classification task would classify this claim as incorrect since there are \textit{50} states in the United States. However, this claim actually contains one valid sub-claim: ``{\em United States is in \underline{North America}}'' \textcolor{green}{\ding{51}}, and one false sub-claim: ``{\em United States has \underline{51} States} \textcolor{red}{\ding{55}}. Providing a single label stating that this claim is not supported or a single score indicating its factual accuracy is not helpful. It would still require users to manually identify text spans corresponding to potential incorrect facts, generate search queries to gather evidence from the open web, and ultimately make a decision based on multiple pieces of evidence.

\highlight{In this work, we present FLEEK (\textbf{F}actua\textbf{L} \textbf{E}rror detection and correction with \textbf{E}vidence Retrieved from external \textbf{K}nowledge), an intelligent and model-agnostic tool designed to support end users (e.g. human graders) in fact verification and correction.} Our tool features an intuitive and user-friendly interface, capable of automatically identifying potential verifiable facts from input text. It generated questions for each fact and queries both curated knowledge graphs and the open web to collect evidence.
Our tool then verifies the correctness of the facts using the gathered evidence and suggests revisions to the original text. 

\begin{figure}[th!]
    \centering
    \begin{subfigure}{0.32\textwidth}
        \centering
        \fbox{\includegraphics[width=\textwidth]{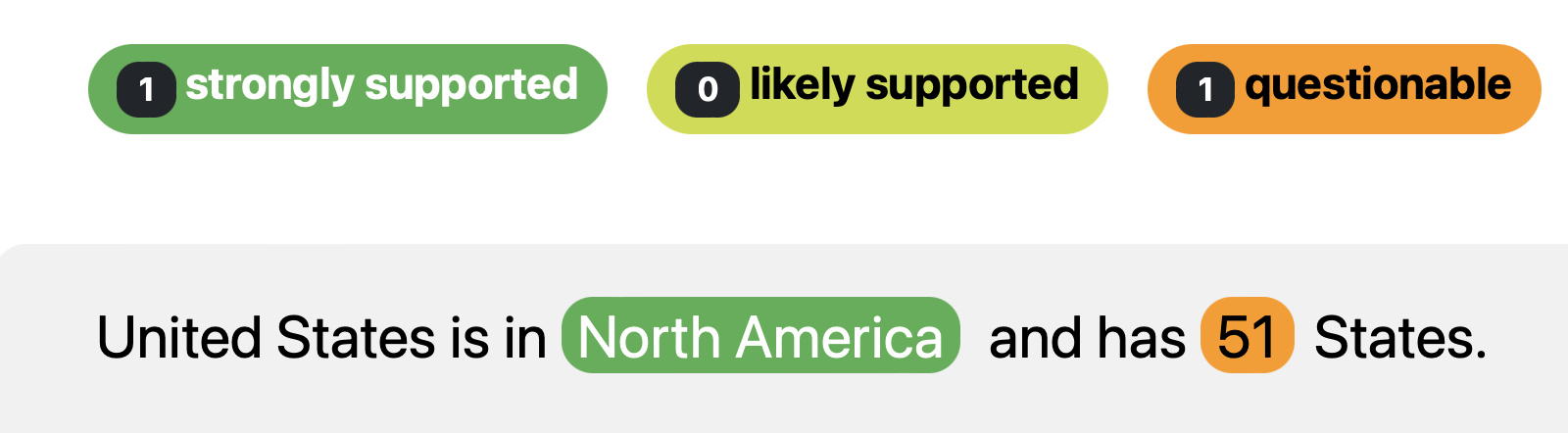}}
        \caption{Factuality annotations by \tool}
    \end{subfigure}
    \hfill
    \begin{subfigure}{0.32\textwidth}
        \centering
        \fbox{\includegraphics[width=\textwidth]{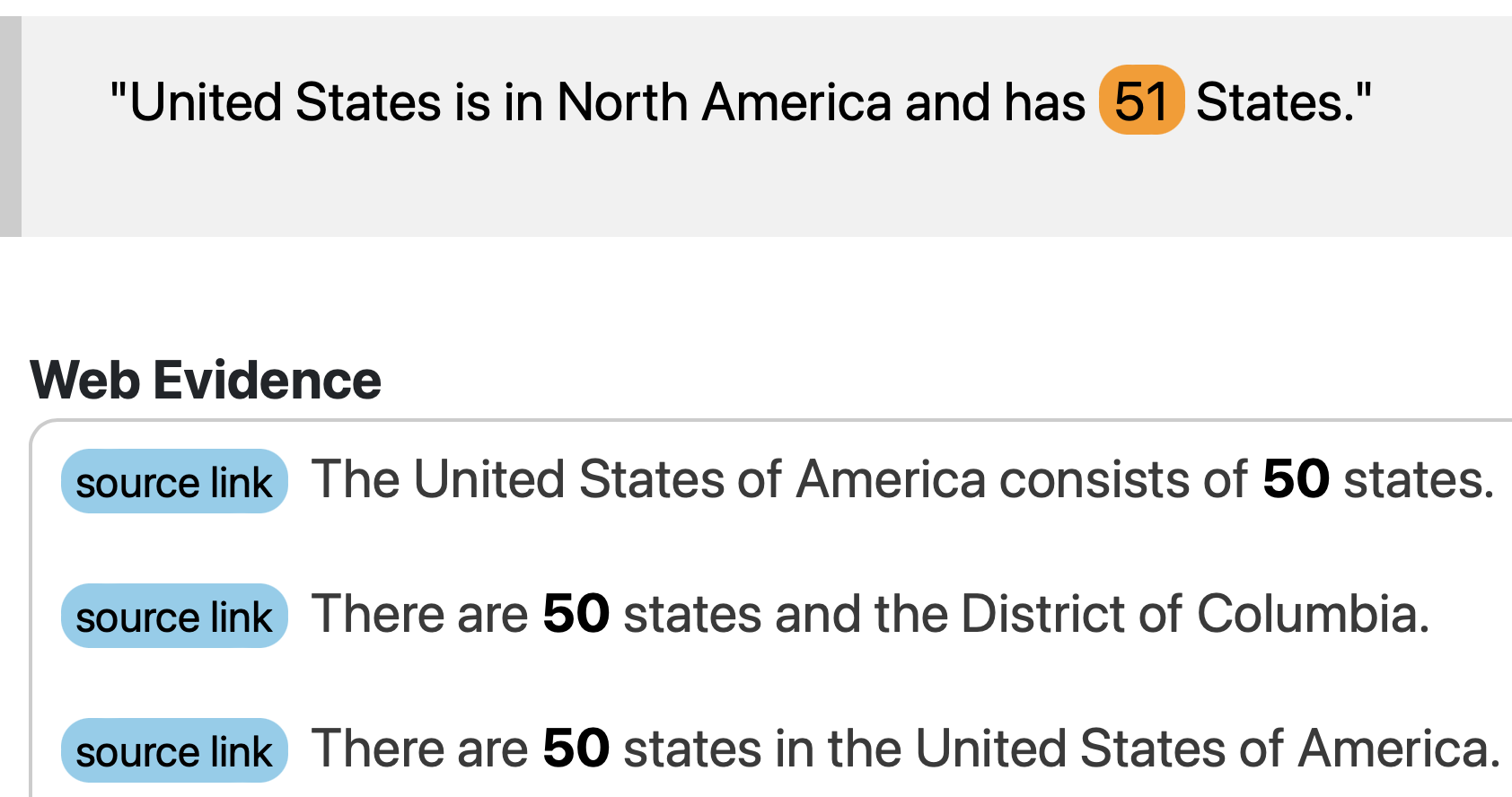}}
        \label{fig:subfigure2}
        \vspace{-4mm}
        \caption{A clickable questionable fact}
    \end{subfigure}
    \vspace{-3mm}
    \caption{Screenshots of \tool}
    \vspace{-3mm}
    \label{fig:motivating_example}
\end{figure}

Our verification process is naturally interpretable since the extracted facts, generated questions, and retrieved evidence all directly reflect which information units contribute to the verification process. For the example mentioned above, \tool would highlight verifiable facts with different colors indicating their factuality levels (see Figure \ref{fig:motivating_example}(a)), and these clickable highlights can open a dialog that further lists evidence retrieved to support or refute each claim (see Figure \ref{fig:motivating_example}(b)).

To the best of our knowledge, \tool is the first verification and correction system that provides fact-level decisions, attributes them with evidence from online sources of information, and proposes factual revisions.

\section{Methodology}
\begin{figure}[h]
    \centering
    \includegraphics[scale=0.42]{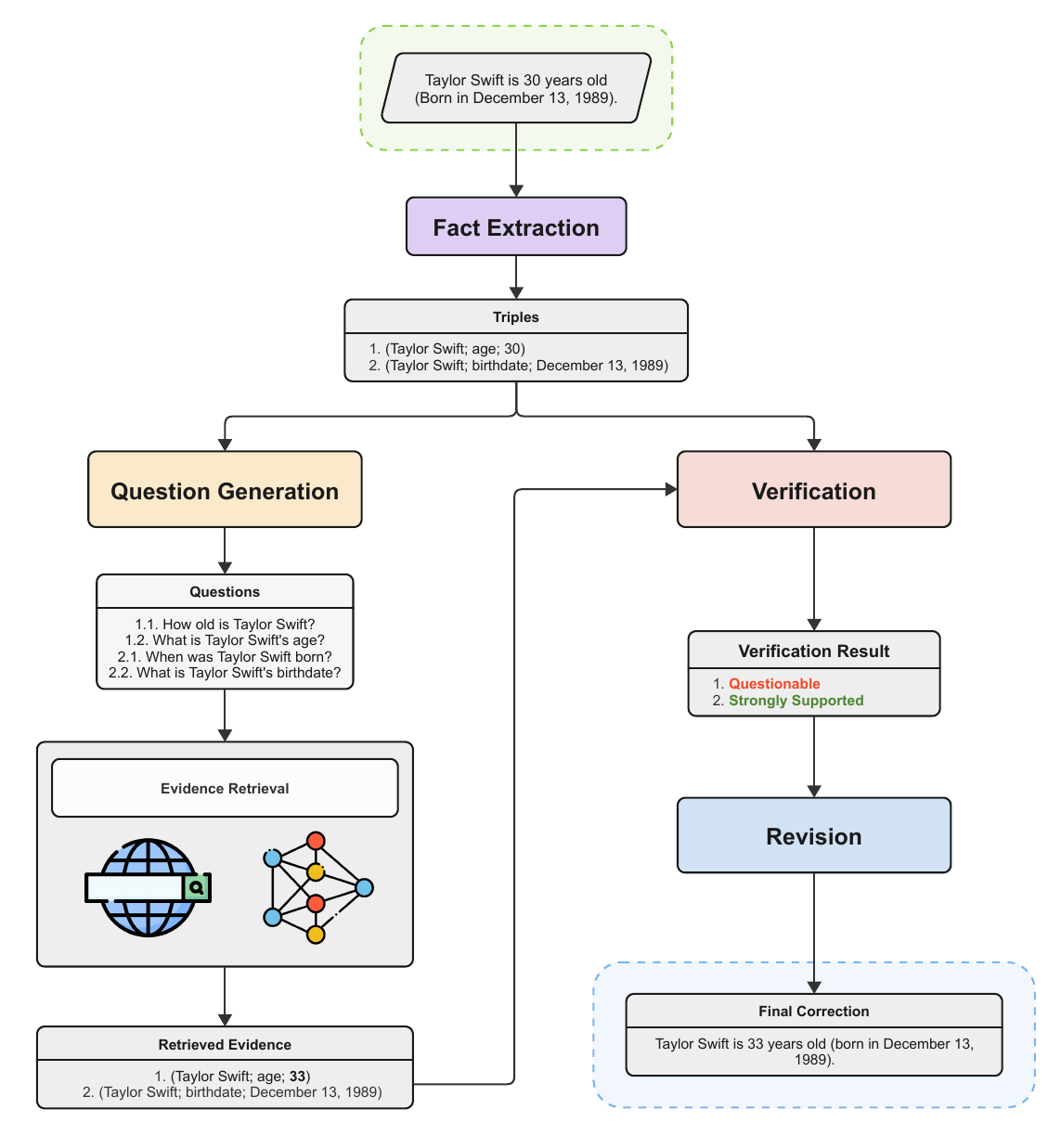}
    \vspace{-8mm}
    \caption{\tool verification and revision framework.}
    \label{fig:approach}
    \vspace{-5mm}
\end{figure}

Figure \ref{fig:approach} shows the overall architecture of \tool. Basically, \tool can perform two tasks: \textit{Fact verification} and \textit{Fact Revision}. 
Next, we describe the methodology used to enable the two tasks. 
\ignore{
An overview of our system is shown in Figure \ref{fig:approach}. We break the fact verification task into 4 sub-tasks: (1) triple extraction: get the potentially verifiable sub-claims; (2) question generation: create queries to retrieve relevant evidence from different sources; (3) evidence retrieval: collect evidence from highly accurate sources (e.g., curated KG) and open web to compensate the low coverage of KG \cite{10.1145/2623330.2623623, peng2023knowledge}; (4) reasoning: generate factuality scores for the verifiable facts. Finally, \tool also suggests revisions to improve the factuality of the original claim. We next describe these tasks in more detail. }

\subsection{Fact Verification} \label{sec:fv}
Given an input passage $p$, we split it into a set of sentences $\{s_1, ... s_i\}$. We then verify each sentence using the sequential pipeline described below.

\subsubsection{Fact Extraction} \label{sec:fc}
In this work, we define a \textit{fact} as a unit of information that (1) describes a certain entity (2) captures the relation between two entities (3) describes an event. Each fact consists of a subject, a predicate, and at least one object. We use the semi-structured triple format to represent such a fact. Our goal is to break a sentence into a set of triples such that each triple represents a verifiable piece of information. This way, we can provide more fine-grained verification details for each sentence. \ignore{Breaking a sentence into triples is crucial since our goal is to provide more fine-grained verification details since triples in a sentence often lead to verifiable facts.} \ignore{In \tool, we consider two triple formats:}
\camReady{To exhaustively extract facts, we consider two triple formats:}

\noindent
\textbf{Flat Triple}: \camReady{For} binary predicates, i.e., predicates with one object, we represent the fact in the form of \textit{(Subject; Predicate; Object)}. For example, the triple representation of the fact \textit{``Taylor Swift is 30 years old.''} is \textit{(Taylor Swift; age; 30 years old)}. 

\noindent
\textbf{Extended Triple} \citet{Ilyas_2022}:
For $n$-ary predicates where $n>2$, i.e., predicates with multiple objects, we utilize the \textit{extended triple} format to capture the relations between fact constituents. The extended triple format is \textit{(Subject; Predicate; Predicate\_ID; Predicate\_attribute; Object)} where \textit{Predicate\_ID} is an artificial predicate identifier, \textit{Predicate\_attribute} is the name of the predicate's attribute, and \textit{Object} is the attribute's value. For instance, the representation of the sentence \textit{``Taylor Swift moved to Nashville at the age of 14.''} is:

\noindent
\textit{(Taylor Swift; moved; move\_ID; place; Nashville)}

\noindent
\textit{(Taylor Swift; moved; move\_ID; age; 14).} 

\looseness=-1 For an input sentence $s_i$, the task of Fact Extraction is to extract flat triples $T_f = \{t_{f_1}, ..., t_{f_m}\}$ and extended triples $T_e = \{t_{e_1}, ..., t_{e_n}\}$. \camReady{These triples are extracted from the sentence using an open information extraction format, with each triple representing a single predicate attribute.} The final output of this component is: $T = T_f \cup T_e.$

To extract these triples, we came up with five challenging human demonstrations such that, for an input sentence, they include different combinations of flat and extended triples. We prompt two instructable LLMs to obtain such triples. \camReady{More details on LLMs utilized for this task, along with an in-depth analysis of the errors they generate, are provided in section \ref{sec:eval}.}

\ignore{For this initiation of \tool, we prompt \llm to obtain such triples, and \llm can be replaced with any instructable LLMs.}
\ignore{As this is a challenging task that requires entity and semantic relation understanding, we prompt \llm to obtain triples. 
}

\subsubsection{Question Generation (QGen)}
Given the output of the Fact Extraction component $T$, the task of Question Generation is to generate questions for each $t \in T$ such that the answer to the question is the \textit{Object} part of $t$. In this way, various answers retrieved from different sources can be used to verify each triple $t$. Depending on the format of triple $t$ (flat or extended), we introduce two different question generation paradigms.


\noindent
\textbf{Type-aware Question Generation (TQGen)}. Consider the triple \textit{(Taylor Swift; birthdate; 1989)}. If the output of the QGen component is: \textit{``When was Taylor Swift born?''}, the retrieved evidence would probably be the exact birthdate. To generate a question as specific and close to the answer as desired, we propose a type-aware question generation approach. Using TQGen, the generated question will be: \textit{``In which year was Taylor Swift born?''}. To this end, we adopt the \textit{Chain-of-Thought} paradigm. This involves two steps: we instruct our model to first find the \textit{``type''} of the \textit{Object} in the input triple, and then generate a question conditioned on the obtained type information. TQGen guides the subsequent information retrieval component to target and retrieve the exact fact that we aim to verify. Prompting LLMs with two human demonstrations was sufficient for this task.
\ignore{For this instantiation of \tool, we prompt LLMs with three human demonstrations.}


\noindent
\textbf{Context-driven Question Generation (CQGen)}. In addition to generating a precise type-aware question, we need to provide context for extended triples so that the retrieved evidence corresponds to the exact situation that requires verification. Consider the extended triples mentioned earlier, 

\noindent
\textit{(Taylor Swift; moved; move\_ID; place; Nashville)}

\noindent
\textit{(Taylor Swift; moved; move\_ID; age; 14)}

\noindent
where the focus is on generating a question for the first triple. If we only feed the first triple to QGen, the output would not consider the time when the relocation happened. To generate a context-driven question, we need to also feed the second triple, the \textit{context triple}, to Context-driven QGen (CQGen). The output of CQGen in this example is \textit{``To which city did Taylor Swift move to at the age of 14?''}. For this task, we prompt LLMs with two examples.




\begin{figure*}[h]
    \centering
    \includegraphics[scale=0.45]{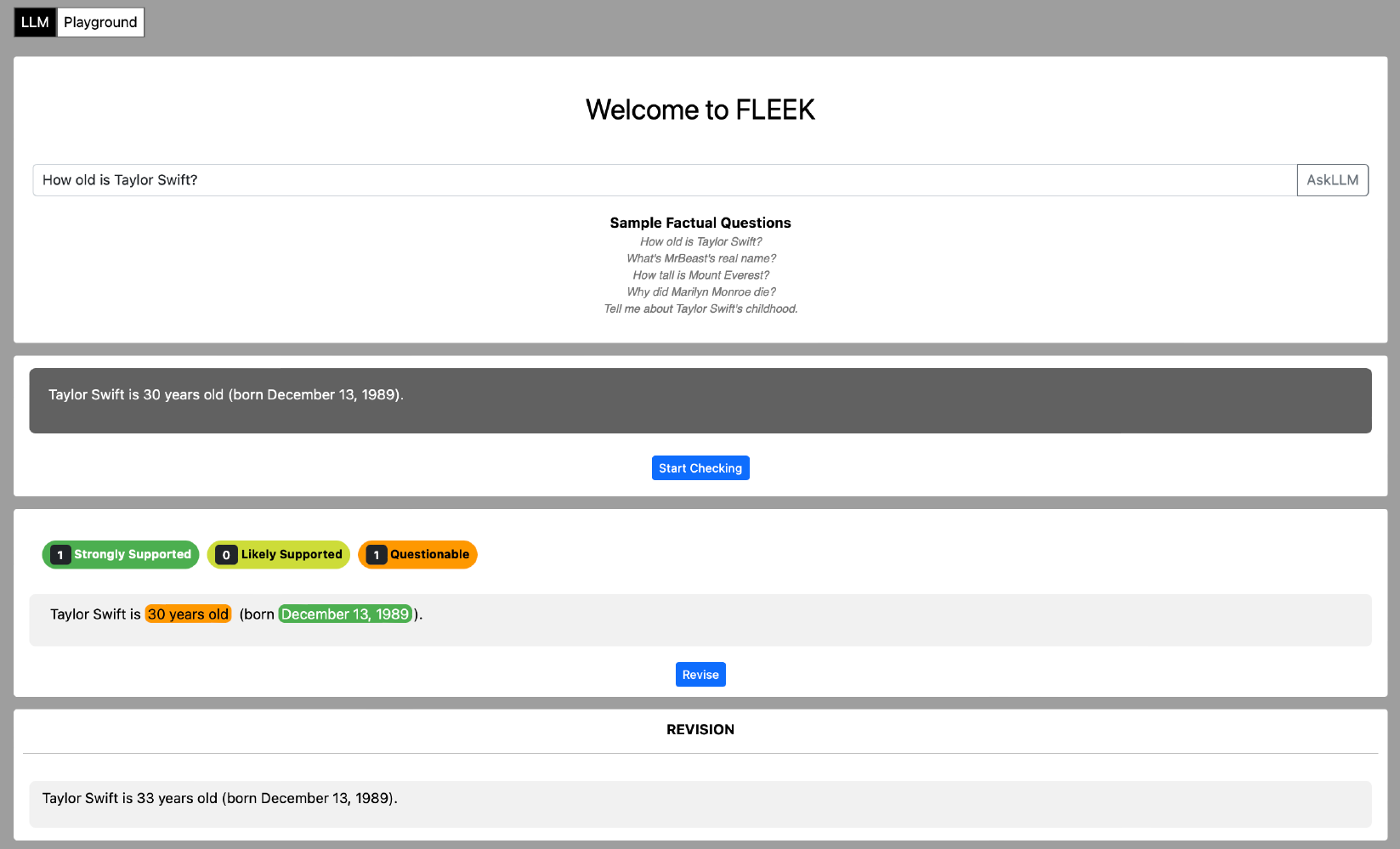}
    \vspace{-1mm}
    \caption{\tool LLM view.}
    \label{fig:llm}
    \vspace{-2mm}
\end{figure*}

\subsubsection{Evidence Retrieval}
The generated questions will be sent to two retrieval systems: a knowledge graph (KG)-based system and a web-based system.

\noindent
\textbf{Knowledge Graph-based}:
We send the question generated for each triple $t$ to our KG question answering (KGQA) system and collect the retrieved short answers. The answer and can either be a single value (e.g., birth date, birthplace) or a list (e.g., profession, spouses). The ensuing entailment decision is derived differently for these two forms of answers (more details in Section \ref{sec:ent}). 

\noindent
\textbf{Web-based}: 
Similarly, we also submit the same question(s) to our web search engine (Web Search). We then take the top-k (e.g., 5) web passages returned for each question and combine them to create a consolidated set of answers. Additionally, Web Search is able to highlight the short answer $a$ for each retrieved passage $p$. The final retrieval list from Web Search is in the format $[(p_1, a_1), (p_2, a_2), ..., (p_k, a_k)]$.   

\begin{figure*}[h]
    \centering
    \includegraphics[scale=0.45]{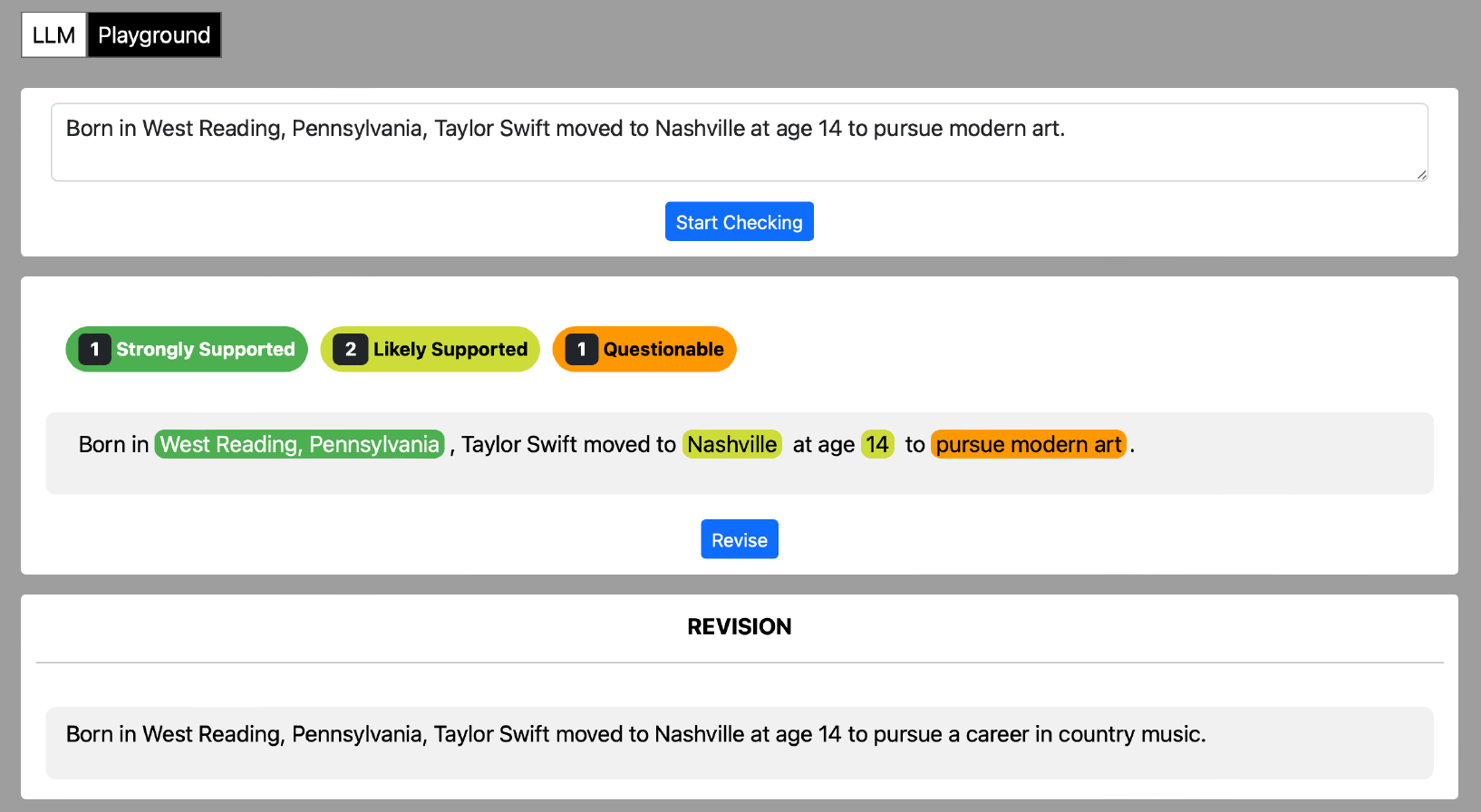}
    \vspace{-1mm}
    \caption{Playground view.}
    \label{fig:playground}
    \vspace{-3mm}
\end{figure*}

\subsubsection{Verification} \label{sec:ent}
Given the triple representation of a fact $t$, the set of KG answers $A_{kg} = \{a_1, a_2,...\}$, and the set of Web answers $A_w = \{(p_1, a_1), (p_2, a_2), ..\}$, the task is to decide whether $t$ is supported by the set of retrieved evidence. This involves two steps:

\noindent
\par{\textbf{Step 1 - Verify against KG answers.}} 
Based on our observation, when the evidence retrieved from the KG is a singular value, the expected answer to the question is most likely to also be a single value (e.g. city of birth). Therefore when $|A_{kg}| = 1$, we classify the fact as \textit{``Strongly Supported''} if it is entailed by the answer, and \textit{``questionable''} otherwise. 
However, if the KG answer is a list, we classify each answer in $A_{kg}$ as either \textit{``supporting''} or \textit{``not supporting''} based on whether it entails the fact. In this case, due to the limited coverage of facts in KG \cite{10.1145/2623330.2623623, peng2023knowledge}, we verify the fact $t$ against web answers as well. 

\noindent
\par{\textbf{Step 2 - Verify against Web answers.}} In case the KG answer is empty or a list, web answers will be also used to make a decision. We classify the answers in $A_{w}$ as either \textit{``supporting''} or \textit{``not supporting''} evidence. Finally, the fact is labeled as \textit{``Likely Supported''} if our system finds at least one \textit{``supporting''} evidence and \textit{``Questionable''} otherwise. In what follows, we describe how to perform evidence classification.

\textbf{Triple Entailment.}
For every triple $t$, we have a set of retrieved answers $A = A_{kg} \cup \{a_i | a_i \in A_w\}$. Our task is to classify each answer as either \textit{``supporting''} or \textit{``not supporting''}. To this end, we construct an evidence triple $t_e$ by replacing the object part of the triple with the short answer retrieved. Therefore, for each $a \in A$ and triple $t = (S; P; O)$, the corresponding evidence triple is $t_e = (S; P; a)$. If the claim triple $t = (S; P; Pid; P\_attr; O)$ is extended, the corresponding evidence triple is $t_e = (S; P; Pid; P\_attr; a)$. The claim and its corresponding evidence triple are then used to form a prompt and fed to LLM to make a final decision. 


\subsection{Fact Revision}
The Fact Revision module aims to correct a questionable fact triple stated in an input sentence into its corrected version while preserving everything else stated in the sentence. More specifically, let $s$ be a sentence containing a questionable triple $t_{src}$, i.e., $s \models t_{src}$ (i.e., $s$ entails fact $t_{src}$). Let the evidence triple formed by the verification process outlined above be $t_{dest}$. The Fact Revision model will thus rewrite $s$ into $s'$ such that $s' \not\models t_{src} \land s'\models t_{dest}$, and $s' \models t_i$ where $t_i\neq t_{src}$ is any triple entailed by $s$. Following is an example (the objects of the triples are in bold):

\noindent
$s = $ \textit{``Taylor Swift is \textbf{30} years old.''}\\
\noindent
$t_{src} = $\textit{(Taylor Swift; age; 30)}\\
\noindent
$t_{dest} = $\textit{(Taylor Swift; age; 33)}\\
\noindent
$s' = $ \textit{``Taylor Swift is \textbf{33} years old.''}

\noindent In our implementation, we prompt LLMs with one demonstration to obtain satisfactory results.

\section{The User Interface of \tool} \label{sec:UI}
\looseness=-1 
\ignore{Our goal is to provide an intuitive yet powerful tool that facilitates users to easily perform fact verification tasks with only a few mouse clicks. To achieve this ambitious goal, we carefully designed the user interface of \tool and provided various ways that users can interact with it. }
The frontend of \tool is built using Angular\footnote{\url{https://angular.io/}} and Bootstrap UI\footnote{\url{https://getbootstrap.com/}}, which allows for creating dynamic, interactive, and visually appealing user interface. The backend of \tool is handled by Django\footnote{\url{https://www.djangoproject.com/}}, a Python-based server-side framework that facilitates the integration with ML-based libraries. The entry point to the system is the two views, {\tt LLM} and {\tt Playground}, shown at Figure \ref{fig:llm}, which we describe next. 

\noindent{\bf LLM View}.
In this view, the user can check the factual consistency of an LLM (e.g. GPT-3.5) that the user provided (as an endpoint). To interact with \tool, first in the \textbf{Input Panel} (Figure \ref{fig:llm}, upper panel), the user can type their query in the question bar, e.g., `{\tt How old is Taylor Swift?}' (or click one of the sample queries) and hit the {\tt AskLLM} button. \tool would send the query to the LLM (\llm in this example) and render its response in the \textbf{Response Panel} (Figure \ref{fig:llm}, the upper second panel in dark grey). The verification process will kick in once the user hit the {\tt Start Checking} button. \tool verifies the claim(s) made by the LLM by going through the process described in Section \ref{sec:fv}. The verification results are shown in the \textbf{Verification Panel} (Figure \ref{fig:llm}, middle panel). 
With our design, \tool is able to highlight the sub-claims in the original text with different color codes to indicate their factual accuracy categories based on the collected evidence. Additionally, the highlighted spans are clickable, which leads to a detailed dialog containing the evidence associated with the claims (illustrated in Figure \ref{fig:motivating_example} and Figure \ref{fig:desc_llm}). Evidence retrieved from the web is accompanied by a {\tt source link} as well. 
At the bottom, the user can request \tool (hit the {\tt Revise} button) to revise the original claims using the collected evidence. Based on the evidence retrieved from the KG and the web, we can have multiple revision alternatives. Verification results for the example shown in Figure \ref{fig:llm}  and \ref{fig:playground} allow for only one possible revision.

\noindent{\bf Playground View}.
This view allows the user to verify any specific piece of text of their choosing. This feature empowers users to automatically fact-check tweets, trending news, arbitrary LLM outputs, or even their own writing with just a few clicks. Figure \ref{fig:playground} illustrates the view.
The user can input their desired text into the designated input panel (scratchpad) and hit the \textit{"Start Checking"} Button (Figure \ref{fig:playground}, upper panel). The verification and revision process is the same as in the \textit{LLM View}.


\begin{table*}[htb]
\centering
\small
\begin{tabular}{|c|c|ccccccc|} 
\hline
\multirow{2}{*}{$\textbf{Instance}$} & \multirow{2}{*}{$\textbf{Category}$} & \multicolumn{3}{c}{\textbf{$Bench_{LLM}$}} && \multicolumn{3}{c|}{\textbf{$Bench_{Text}$}} \\
\cline{3-5} \cline{7-9}
& & P & R & F1 & & P & R & F1\\
\hline
\multirow{4}{*}{$\tool_{Vicuna}$} & Strongly Supported & 91.66 & 84.61 & 87.99 & & 84.61 & 91.66 & 87.99 \\ 
& Likely Supported & 94.73 & 58.69 & 72.47 & & 87.5 & 37.33 & 52.33 \\
& Questionable & 54.54 & 60 & 57.13 & & 93.61 & 74.57 & 83.01 \\
\cline{2-9}
& Total & 88.75 & 61.73 & \textbf{72.81} & & 90.21 & 56.84 & \textbf{69.73} \\
\hline
\multirow{4}{*}{$\tool_{GPT-3}$} & Strongly Supported & 100 & 95.23 & 97.55 & & 100 & 100 & 100 \\ 
& Likely Supported & 93.22 & 61.79 & 74.31 & & 95.77 & 79.06 & 86.61 \\
& Questionable & 66.66 & 100 & 79.99 & & 87.75 & 69.35 & 77.47 \\
\cline{2-9}
& Total & 89.0 & 76.06 & \textbf{82.02} & & 93.28 & 77.16 & \textbf{84.45} \\
\hline
\end{tabular}
\vspace{-2mm}
\caption{\label{main_results} Evaluating two instances of \tool on Bench$_{LLM}$ and Bench$_{Text}$.}
\vspace{-4mm}
\end{table*}

\section{Evaluation} \label{sec:eval}

Previous benchmarks on fact verification \cite{Thorne18Fever, Aly21Feverous} provide a single decision for the entire claim based on the retrieved evidence. However, in this work, we introduce fine-grained fact verification with attribution to external knowledge. As this is the first study on this task, there exist no benchmarks for evaluating \tool's performance. Next, we conduct preliminary experiments using manually created evaluation data.

\subsection{\highlight{Evaluation Data Creation}}\label{sec:bench}
Our system has two use cases. The first one is to verify the responses generated by LLMs (in this case, \llm). To evaluate our system's performance, we selected 50 questions from WikiQA (Wikipedia open-domain Question Answering) test set \cite{yang-etal-2015-wikiqa} and collected their corresponding \llm responses. We then manually annotated each response using the following steps: (1) identify the facts within the response, (2) label each fact as \textit{``Strongly Supported''}, \textit{``Likely Supported''}, or \textit{``Questionable''}, (3) accompany each fact with an evidence set, particularly the questionable facts. \highlight{We call this dataset \textit{Bench$_{LLM}$}. Each instance in the \textit{Bench$_{LLM}$} contains the annotated GPT-3-generated response. }

The second use case is to verify an arbitrary input text. To create evaluation data that suits this task, we target the introduction section of Wikipedia pages. To partially perturb sentences and create incorrect facts, we sample 50 random sentences with at least one hyperlink. Then, we retrieve the hyperlink's corresponding entity from Wikidata \footnote{\url{https://wikidata.org/}}, find the entity's type (\textit{instance of} property), and retrieve candidate entities with the same type. Finally, we perturb the sentence by replacing the original hyperlink with one randomly selected entity within the candidate list. After perturbation, we annotate the sentence the same way that we created \textit{Bench$_{LLM}$}. \highlight{We call this dataset \textit{Bench$_{Text}$}.}

\subsection{Large Language Models}
All \tool's components that facilitate fact verification and correction use few-shot prompting with a large language model. Any model that can learn from in-context demonstrations can be used to instantiate \tool. We chose one open-source model, Vicuna (33 billion parameters), and one closed-source model, \llm (175 billion parameters), to create two instances of our tool. We call the instance with Vicuna as its large language model \tool$_{Vicuna}$ and the instance that utilizes \llm as its large language model \tool$_{GPT-3}$. We evaluate both instances in the following section. 

\subsection{Experimental Results}
Consider the set of system-generated spans $S = \{s_1, ..., s_n\}$ and ground truth spans $G = \{g_1, ..., g_m\}$. We measure the number of textual spans that are correctly identified, labeled, and attributed to the valid supporting evidence as $ov$. Then, we calculate verification system's precision as $\frac{ov}{|S|}$, recall as $\frac{ov}{|G|}$, and the F1 score. Table \ref{main_results} shows the performance of our verification systems on both evaluation datasets. As illustrated, \tool$_{GPT-3}$ outperforms \tool$_{Vicuna}$ on both datasets by a large margin ($12$ F1 pts). However, given that Vicuna is about $5\times$ smaller than \llm, the average performance of \tool$_{Vicuna}$ ($71.27$ F1) shows its efficiency in Fact Verification. 
\noindent
Moreover, the results show that both systems can identify the \textit{``Strongly Supported''} facts with high precision and recall. However, they fail to detect all facts or attribute them to the correct evidence for \textit{``Likely Supported''} or \textit{``Questionable''} cases. 


We also measure the accuracy of revisions proposed by the fact correction component. Both systems have on-par performance with an average accuracy of 72.7\%. However, our investigation shows that 54.1\% of incorrect revisions are a result of errors in previous components propagated through the system. Thus, Fact Correction's average precision, given the correct verification results, is 87.5\%.

Note that although our initial results show great promise, both evaluation datasets are small (50 sentences) and come from the same data source (Wikpedia). One ongoing work is to create a larger benchmark (with different levels of difficulty from more diverse sources) for a more extensive and reliable evaluation of our system.

\begin{figure}
    \centering
    \includegraphics[scale=0.125]{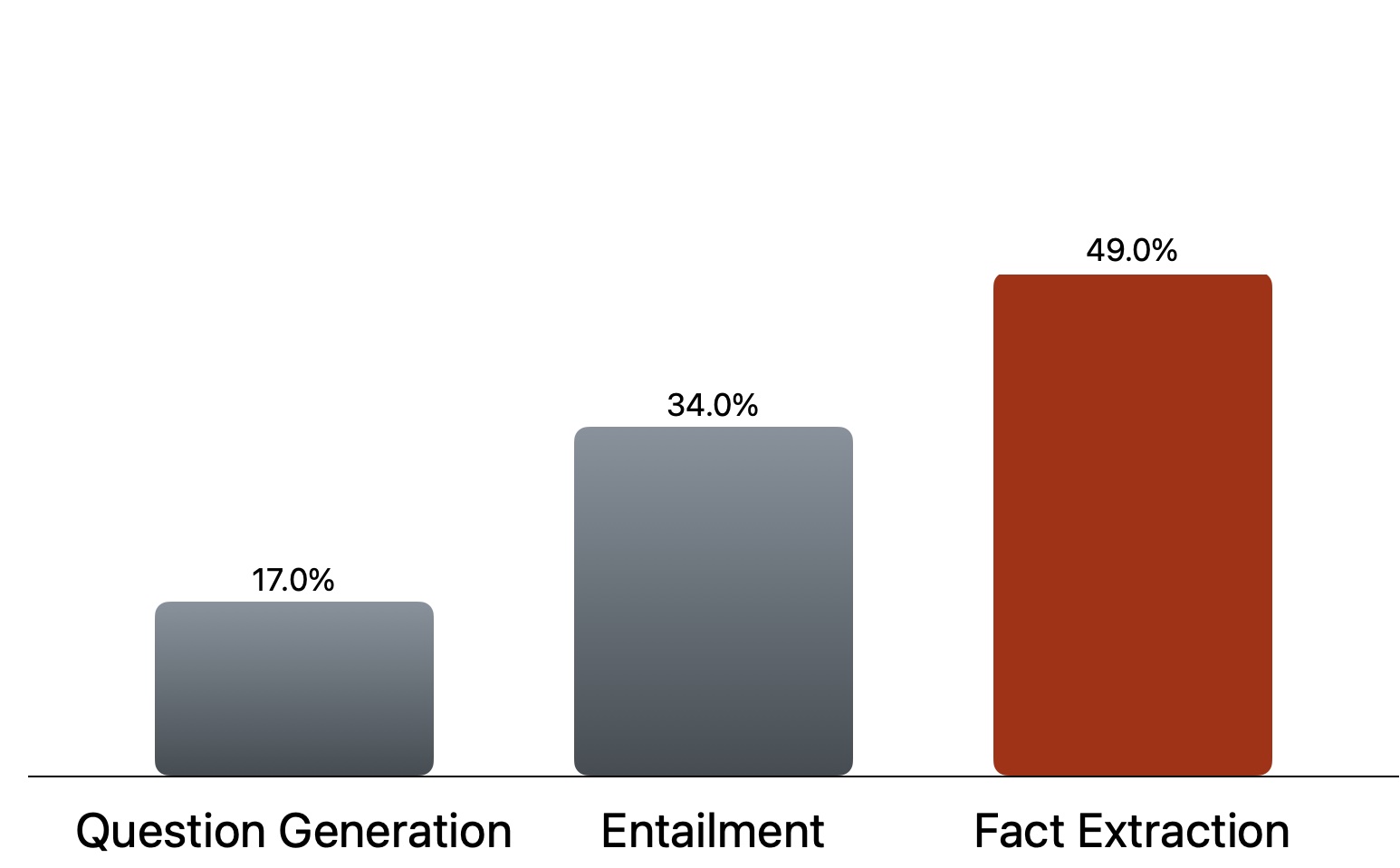}
    \vspace{-3mm}
    \caption{Percentage of total errors generated by different components of \tool$_{GPT-3}$.}
    \label{fig:error_analysis}
    \vspace{-4mm}
\end{figure}

\smallskip
\noindent{\bf Error Analysis.} We randomly select 30 examples where \tool$_{GPT-3}$ made erroneous decisions and investigate the types of errors each of its components made (Figure \ref{fig:error_analysis}). In general, the Fact Extraction component accounted for a significant portion, approximately $49\%$, of the total errors. This emphasizes the difficulty of mastering Fact Extraction through in-context learning. Errors produced by this component include, but are not limited to, wrong triple format, broken n-ary relations, missing triples, and hallucination. Figure \ref{fig:error_analysis} further indicates that the \llm might not excel in reasoning, as the entailment component also contributes significantly to system errors.

\vspace{5mm}
\section{Conclusion and Future Work}
We presented \tool, an innovative solution geared towards assisting users in verifying the accuracy and factuality of textual claims. We aim to keep improving the \tool so that it can be a handy tool for various stakeholders. As part of our future work, we intend to do more comprehensive evaluations of \tool, including testing it with various LLMs and over a comprehensive benchmark.

{\bf Limitation.}
First, our current system depends on the initial set of responses generated by LLMs to perform the tasks. Nevertheless, we can prompt each component multiple times and employ methods such as majority voting to enhance the accuracy of each task. Second, the experiments presented are based on small-scale datasets. We plan to expand both datasets as part of our future endeavors. Finally, both datasets are manually annotated by one annotator. We plan to hire more annotators and refine the annotation process so as to provide a more comprehensive evaluation of our method.


\bibliography{anthology,custom}
\bibliographystyle{acl_natbib}

\appendix



\end{document}